# Open challenges in understanding development and evolution of speech forms: the roles of embodied self-organization, motivation and active exploration[1]

Pierre-Yves Oudeyer[2]
Inria, France
Ensta ParisTech, France

**Abstract**: This article discusses open scientific challenges for understanding development and evolution of speech forms, as a commentary to Moulin-Frier et al. (Moulin-Frier et al., in press). Based on the analysis of mathematical models of the origins of speech forms, with a focus on their assumptions, we study the fundamental question of how speech can be formed out of non-speech, at both developmental and evolutionary scales. In particular, we emphasize the importance of embodied self-organization, as well as the role of mechanisms of motivation and active curiosity-driven exploration in speech formation. Finally, we discuss an evolutionary-developmental perspective of the origins of speech.

## 1. Comparing theories of speech formation in a unified Bayesian framework

Studying the forms and formation of speech has long been a topic of tremendous interest for cognitive science in general. It has been repeatedly used in the last century as the cradle in which alternative theories of language as well as sensorimotor control have been expressed and debated. Jakobson (Jakobson, 1941) used it as a strong ground for the early elaboration of structuralist theories of cognition. Later on, it has been the pivot of theories of perception, and their potential links to action (Galantucci et al., 2006), as well as theories of language development in the child (Oller et al., 2013). It has also gathered efforts in the quest for understanding the origins of language, where a mystery is how linguistic forms can arise, be shared and evolve in a population of individuals (Steels, 2011; Oudeyer, 2006; Kirby et al., 2014; Moulin-Frier et al., in press).

Across these scientific enterprises, mathematical and computational modeling has been prominently used in the latest decades, grounded in the physics of the speech system and in the dynamics of neural and learning architectures. Such models constitute a formal language allowing us to formulate and analyze precisely hypotheses about complex mechanisms. Yet, an obstacle to scientific progress has been that alternative theories have often been expressed through different formal languages, making it challenging to articulate and compare them in a single framework. This challenge applies both to models of speech evolution (e.g. Liljencrants and Lindblom, 1972; Berrah et al., 1996; Browman and Goldstein, 2000; de Boer, 2000; Oudeyer, 2005; Pierrehumbert, 2006; Wedel, 2011) and models of speech acquisition (e.g. Guenther,

---

1994; Warlaumont et al., 2013; Howard and Messum, 2014; Moulin-Frier et al., 2014). In this perspective, the COSMO Bayesian framework (Moulin-Frier et al., in press) is a significant step forward as it leverages Bayesian modeling to develop a formal framework that allows us to formulate in a unified manner many key theories of speech (Moulin-Frier et al., 2012), as well as theories of how speech forms can arise in populations of individuals.

A particularly useful feature of such a Bayesian framework is that it constrains model designers to be as explicit as possible on the assumptions of their models. Furthermore, Bayesian modeling allows compact expression of relationships between subparts of a model, abstracting details of implementation to highlight the global functional dynamics. In their article, Moulin-Frier et al. show in detail how various theories of speech perception and production compare in the context of communication loops among individuals. They also show how such a framework can encode the dynamics of so-called "language games" to account for the formation of shared speech codes in a population of individuals, as well as to explain why certain vowel and consonant structures are more frequent than others in world languages.

## 2. Speech from non-speech: open questions

As such formalism provides a compact and general view on a large family of models of the formation and learning of speech structures, it also affords identifying open scientific questions. This can be done in particular through analyzing the assumptions of the COSMO framework. As Moulin-Frier et al. very clearly state, COSMO attempts to explain some properties of phonological systems out of speech communication principle, i.e. out of (Bayesian) optimization processes that lead individuals to learn and negotiate a communication system that is efficient under physiological constraints. Furthermore, thanks to the method of Bayesian modeling, Moulin-Frier et al. also make it explicit that their model assumes mechanisms that solve requirements of "adequacy" (availability of a system of forms easy to produce and perceive), "parity" (capability to play symmetric roles in speech interaction) and "reference" (capability to use a device like pointing to ensure shared attention on a referent).

Thus, the model relies on a pre-existing set of linguistic abilities, as well as abstracts away from many non-linguistic processes, such as sensorimotor development outside speech, non-linguistic activities such as sensorimotor coordination in joint tasks, or properties of the body outside the speech system.

This in itself is not a weakness of the framework, especially because this is made explicit. But it points to a very important question, formulated already long ago by Lindblom (Lindblom, 1984): "[Which are the processes that allow to] derive language from non language"?. This question applies at several scales: development in individuals, cultural evolution and phylogenetic evolution in populations.

At the developmental scale, one needs to explain how young infants come to discover and master speech sounds, and how they understand that these sounds can be used to produce effects on their social peers and coordinate with them. Infants indeed are not born with a refined understanding of what "speech communication" is, and optimization processes driving the formation of (speech) codes efficient for communication may hardly be at work at the beginning. Indeed, notions of "code" and "communication" shall themselves be formed though cognitive and social development, leveraging in particular the capability to assign new functions to behaviors previously mastered (which has been called "functional flexibility", Oller et al., 2013).

At the evolutionary scales, cultural and phylogenetic, an analogous mystery is still far to be solved: how did the capacity to linguistically communicate through speech or gestures appear out of non-language? Language games models such those presented in Moulin-Frier et al. have mostly focused on the question of how a shared linguistic convention can form and change at the population level, but assuming the capacity to handle the syntax and meaning of these language games, i.e. assuming a capacity for language. But how did language form in communities of individuals who did not already have such tools to build and negotiate a linguistic system?

Speech, and language in general, are embedded in a network of diverse non-linguistic activities, as well as influenced by constraints due to biological implementation of the body and the brain: what consequences can this have on the formation of speech forms at developmental and evolutionary scales?

Such non-speech mechanisms from which speech communication shall emerge, or be influenced by, are bound to have consequences both at the individual/developmental level and at the population/evolutionary level, acting as structure providers and filters constraining and guiding the formation of speech forms.

We will now discuss three families of non-speech mechanisms that may be useful starting points to further understand how speech can be formed out of non-speech: self-organization and spontaneous pattern formation in physical systems, the role of intrinsic and extrinsic motivational systems, and finally some commonalities between speech development and the development of other sensorimotor skills through the prism of active exploration.

**2.1 Embodied self-organization of speech forms**

Nature is full of complex organized patterns, in particular in the inorganic world: spiral galaxies, sand dunes, deltas of rivers, polyhedrons in water foam, ice crystals are all macro-patterns that spontaneously form out of the physical interaction of their micro-components. Such self-organization of structures, due to the laws of physics in complex systems, is also at play in the living world. For example, it has been identified in the formation of spots and stripes on the skin of animals, of hexagonal honeycombs, or for organizing group behavior in insects or birds (Ball, 2001). At the developmental scale, these spontaneous patterns can be at play to generate organized behavior without any process of explicit optimization. At the evolutionary scale, such self-organized developmental processes can act as constraints, or might have been recruited and shaped to serve optimally a functional purpose.

Let us take the example of biped locomotion. Walking implies the real-time coordination of many body parts. Each of our bones and each of our muscles are like the musicians of a symphonic orchestra: they must produce a movement impulse (or silence) at the right moment; and it is the juxtaposition and integration of all these impulses and silences which builds the symphony of the whole body walking forward with elegance and robustness. But is there a musical score which plans these coordination details? Is there a mechanism in the brain that, every few milliseconds, observes the current state of the body and environment and computes the optimal muscular activations to maintain balance and move forward with minimal energy consumption? This is the hypothesis pursued in several strands of research, for example the theory of optimal motor control in humans and robots (Todorov and Jordan, 2002).

However, experiments on passive dynamic walking have shown that explicit optimization of balance and energy consumption may not be the full story to account for the structure of biped gaits. For example, Tad McGeer built a pair of mechanical legs, without a motor and without a computer (thus without the possibility to make calculations), and reproduced the geometry of human legs (McGeer, 1990). Then, he threw the robot on a little slope, and the robot walked: automatically, through the physical interaction between the various mechanical parts and gravity, the two legs generated a gait that looked surprisingly similar to a human gait, and was robust to disturbances. Other laboratories replicated the experiment many times (Collins et al., 2005).

The vocal tract and its motor system constitute also a complex physical system with coupled dynamics. Are there structures of speech which, like passive dynamic gaits, form spontaneously out of the physics of the vocal tract, already providing a highly constrained space of forms in which speech communication principles can carve signals? A theoretical exploration in this direction has been the dynamical systems approach to speech motor control elaborated by Kelso et al. (Kelso et al., 1986). Such a perspective partially questions the scope of modeling approaches that aim to predict precisely the forms of speech without relying on a detailed model of the physics of the vocal production system.

Coupled mechanical systems are not the only potential source of pattern formation which stands out of an optimal control perspective. The neural system may also have intrinsic properties, not necessarily specific to a modality like speech, guiding the formation of structures outside optimization. For example, recent work on human motor control of muscle synergies in the arm have shown that the brain may prefer to reuse good-enough synergies to solve tasks rather than find optimal solutions (De Rugy et al., 2012; Loeb, 2012). To what extent could this apply to speech communication systems? An example comes from a model of the formation of speech sounds in populations of individuals presented in (Oudeyer, 2006). In this model, individuals were equipped with perceptuo-motor neural maps connecting a vocal tract model and an auditory model. These neural maps were composed of neurons which have (initially random) spatiotemporal receptive fields and mechanisms of cellular death under low activation. Random spontaneous activations of these maps lead each individual to produce vocal babbling movements, to learn the association with their auditory consequences, and to stimulate the auditory system of the neighbouring individuals. Experiments showed that these maps spontaneously self-organized combinatorial speech forms shared across individuals of the same group. If an individual was alone with no auditory stimulation from others, self-babbling also led to combinatorial vocalizations. Furthermore, these speech forms were characterized by a phonotactic organization encoding systematic and constrained possibilities of sound combinations that matched coarse tendencies of human languages. Yet, no mechanisms optimizing for speech communication was assumed: the resulting organized speech forms were rather the collateral effect of the natural dynamics of the coupled neural maps in interaction with the morphological properties of the vocal tract (Oudeyer, 2005; 2006;).

A significant challenge thus relies in how we can reconcile (or not) models of the formation of organized vocal structures that do not rely on the optimization of speech communication principles, and models that target to explain properties of speech as an optimal system for transmitting signals efficiently over a physical system. An open question is also whether the Bayesian approaches, which aim to abstract the physical implementation of biological processes, provide the adequate language to account for pattern formation arising because of details of the physical and biological substrate which may a priori be unrelated to the functional structures to be explained.

## 2.2 Intrinsic and extrinsic motivations in speech development

In models of language games (Loreto and Steels, 2007; Steels, 2012), for example in the deictic games of the COSMO framework (Moulin-Frier et al., under review), interactions among individuals happen by pairs: in each interaction episode, two individuals "decide" to choose a topic and exchange information about which signals they use to name it. Individuals are pre-programmed to "want" to communicate with each other. In other models of speech formation at the population level (Berrah et al., 1996; de Boer, 2000; Oudeyer, 2006), or in some models of the acquisition of an existing speech system (Guenther, 1994; Howard and Messum, 2014), one assumes mechanisms which push individuals to systematically produce vocalizations through babbling or be responsive to social feedback within a linguistic perspective. Individuals are here pre-programmed to "want" to practice their speech skills.

But what is the origins of such motivational mechanisms? Are they specific and ad hoc to speech and language, or are they the result of a more fundamental developmental process? Interestingly, observations of infant vocal behavior show that infants explore sound production even in the absence of peers (e.g. in bed babbling) and before they have been flexibly linked to the function of speech communication. Is this a form of play that was specifically selected by evolution to prepare the individual to later language development? Or is this an instance of a more general form of play and exploration?

Recent work on modeling intrinsic motivational systems has suggested hypotheses regarding this question. Research in psychology and neuroscience has identified that our brains have an intrinsic motivation to explore novel activities for the sake of learning and practicing (Lowenstein, 1994; Gottlieb et al., 2013). Neuroscience is beginning to identify brain circuits involved in spontaneous exploratory behaviors and curiosity-driven learning (Gottlieb et al., 2013). A fruitful line of computational models has been considering intrinsically motivated exploration as being driven by the search of learning progress niches (Schmidhuber, 1991; Oudeyer et al., 2007; Oudeyer and Smith, in press): the learner is viewed as a little scientist trying to understand its own body and its relations with the environment through actively selecting experiments which improve the quality of its predictive model, i.e. which provide maximal information gain. These active learning mechanisms have been applied to understanding the exploration of various kinds of sensorimotor spaces, ranging from arm reaching, locomotion and object manipulation (Baldassarre and Mirolli, 2013; Baranes and Oudeyer, 2013; Nguyen and Oudeyer, 2013; Ivaldi et al., 2014).

Focusing on vocal development, Moulin-Frier et al. conducted experiments where a robot explores the control of a physical model of the vocal tract in interaction with vocal peers, driven by an intrinsic motivation to improve its predictions and mastery of its own body (Moulin-Frier et al., 2014). The robot explores the relation between vocal tract movements and the corresponding auditory effect driven actively by an intrinsic motivation to improve its model of the world. Experiments showed how such a mechanism can explain the adaptive transition from vocal self-exploration with little influence from the speech environment, to a later stage where vocal exploration becomes influenced by vocalizations of peers, as observed in human infants (Oller, 2000). Within the initial self-exploration phase, a sequence of vocal production stages self-organizes, and shares properties with infant data: the vocal learner first discovers how to control phonation, then focuses on vocal variations of unarticulated sounds, and finally automatically discovers and focuses on babbling with articulated proto-syllables. As the vocal learner becomes more proficient at producing complex sounds, imitating vocalizations of peers starts to provide high learning progress explaining an automatic shift from self-exploration to vocal imitation.

Thus, in such a model speech structures (up to proto-syllables) develop as the result of a form of curiosity-driven exploration that is not yet connected to the function of speech communication. This is an optimization process (improvement of a predictive model is maximized), but such optimization is not driven by a speech communication purpose.

Another line of work, studying the role of emotional social reinforcement, has explored how other pre-speech mechanisms may help build the ground for speech communication forms. Oller et al. (Oller et al., 2013) has for example discussed in depth the functional flexibility of early speech sounds, and in particular how they can initially be bootstrapped through an extrinsic motivation to share and express emotions with social peers. In a computational model, Warlaumont (Warlaumont et al., 2013) showed how non-linguistic social reinforcement could progressively drive a vocal learner to produce syllable-like vocalizations. Howard and Messum (Howard and Messum, 2014) complementarily explored how such social reinforcement could lead a vocal learner to acquire and match adult speech forms.

These lines of work suggest that intrinsic and extrinsic forms of motivations that are not specifically geared towards speech communication may play early on an important role in carving forms of vocalizations that transform later on into speech (Oudeyer and Smith, in press). A major open challenge appears to understand and model a full account of the transition from these non-speech systems to speech communication systems. For example, an open question is to understand how an intrinsically motivated learner that discovers speech sounds through curiosity-driven exploration and/or through a social process for sharing emotions, could discover that these sounds can also be used as tools to manipulate others and coordinate with them, and how they can be shaped and negotiated through a cultural evolution process alike the COSMO model.

## 2.3 Active exploration: links with the development of other sensorimotor systems

From a sensorimotor learning point of view, developing the skills to produce controlled sounds with a vocal tract shares very similar challenges with learning other skills like arm reaching, legged locomotion or object manipulation. Indeed, learning in all these sensorimotor spaces is difficult because 1) these spaces are high-dimensional and non-linear; 2) learning happens incrementally through physical experiments (e.g. trying a vocal tract movement or an arm movement, observing the produced sounds or hand positions); 3) these physical experiments are costly in time and energy. As a consequence, random exploration of these sensorimotor spaces, i.e. random babbling, is bound to fail, leading to the collection of very sparse sensorimotor observations which cannot be used to infer the regularities of the underlying manifolds (Baranes and Oudeyer, 2013). Thus, exploration needs to be guided.

One could wonder whether there exists guiding mechanisms that are specific to vocal exploration, arm exploration, locomotion exploration, or object manipulation exploration. Maybe there are, beginning with the specific physiological properties of muscle synergies in each modality. But the commonalities between these learning problems strongly suggest a commonality of learning mechanisms, and this is emphasized by computational models of sensorimotor development (Oudeyer et al., 2013). In the previous section, we discussed in particular how intrinsic motivation and social reinforcement mechanisms could guide vocal development. These guiding mechanisms are in fact orthogonal to speech, and have been shown to be highly efficient in guiding the development and acquisition of other sensorimotor skills.

For example, the architecture of curiosity-driven learning used to model speech development in Moulin-Frier et al. (2014) is also a highly efficient active learning method allowing a robot to acquire legged locomotion with a high-dimensional body (Baranes and Oudeyer, 2013), and to learn object manipulation and actively choose who and when to ask help from human teachers (Nguyen and Oudeyer, 2013). As Moulin-Frier et al. (Moulin-Frier et al., 2013) showed that such active exploration mechanisms could self-organize a developmental path where certain speech forms appear with a particular order and timing, similarly the formation of sensorimotor forms in a certain order has been shown in other modalities. In Baranes and Oudeyer (Baranes and Oudeyer, 2013), specific coordination structures leading the robot to walk backwards self-organize before other coordination structure appear for walking forward (a structure also known to appear in human infants). In Oudeyer et al. (Oudeyer et al., 2007; Oudeyer and Smith, in press), the same architecture for active exploration leads a robot to first discover basic affordances between its mouth and objects it can bite, then affordances between its legs and objects it can push or grasp, and finally discover that sounds produced towards another robot provoke predictable reaction in its social peer.

The social guidance mechanisms used by Warlaumont (Warlaumont et al., 2013) or Howard and Messum (Howard and Messum, 2014) to drive the vocal exploration of a learner are also in fact general guiding mechanisms across modalities of sensorimotor learning. Closely related mechanisms have for example shown how a human could use reinforcement signals to progressively shape the exploration and learning of balancing an object (Knox and Stone, 2009), controlling an arm (Pilarski et al., 2011), or combining objects to reach a goal (Thomaz and Brezeal, 2008).

## 3. Towards an evo-devo theory of the origins of speech forms

In the previous sections, we have seen that non-speech mechanisms may play a fundamental role in progressively guiding a young learner into discovering how to use his/her vocal tract to produce speech forms of increasing complexity. Hence, modality-general mechanisms such as active intrinsically motivated exploration and social reinforcement can lead an organism to acquire basic speech forms before he can participate to and understand speech communication per se.

A difficult scientific challenge remains to explain how the child flexibly discovers the *means and goals* of speech communication, i.e. how the child comes to understand and master how certain behaviors he produces (like producing particular speech waves, or gestures if he/she has speech impairments) can be used flexibly and adaptively to get its social peers respond.

Non-speech mechanisms that impact speech development may also be key in understanding how language and languages form at the cultural and evolutionary scale. As we have discussed above, models of speech and language formation in groups of individuals have shown how spontaneous pattern formation in embodied coupled systems could foster the establishment of speech and linguistic conventions. In addition, the mechanisms of intrinsically motivated active learning we discussed also open windows over the evolutionary dynamics of language. Within a context of cultural evolution, an open question is how adult speech forms came to have the structure they have. Language and speech evolve not only through linguistic negotiation between adult peers, but also through cultural transmission to children. And as children acquire the language system of their parents, their learning biases act as filters and modulators over the input from adults. Deacon (Deacon, 1997), and a subsequent series of computational models (Zuidema, 2003; Oudeyer, 2005; Kirby et al., 2014), have shown that cultural processes of language evolution make that linguistic forms adapt not only to become useful tools of communication for adults, but also to be learnable by and "interesting" for infants. Hence, mechanisms of intrinsic motivation which assess "interestingness" in terms of learning progress/learnability (Oudeyer and Kaplan, 2007; Gottlieb et al., 2013) should directly impact what infant will learn and not learn easily, and thus will be key in the cultural evolution of language forms. In addition, mechanisms of active curiosity-driven learning used within the process of language games may significantly improve the speed of convergence towards shared linguistic conventions, as suggested in (Schueller and Oudeyer, 2015).

Beyond understanding the structure of speech forms in human languages, a fundamental question at the cultural and phylogenetic evolutionary scales has been to understand the origins of language itself: how can populations of individual invent and shape the means and goals of language? As argued in Oudeyer and Smith (Oudeyer and Smith, in press), it is important to consider that the non-speech developmental mechanisms that help the child discover speech may also be instrumental in helping populations of individuals to invent language. For example, behavioural innovations resulting from curiosity-driven sensorimotor exploration of what one's body can do to objects and others may provide repertoires of skills, such as organized vocalizations (Moulin-Frier et al., 2014), that could form important elements of a starting kit for language. In this perspective, understanding the interaction between developmental and evolutionary dynamics appear to be a key challenge, within an evo-devo approach (Müller, 2007). However, mathematical and computational models of the formation of speech and language systems at the population level have so far largely abstracted the developmental dimensions. Reversely, models of speech acquisition and development have considered single individuals acquiring an existing speech/language system. Establishing the foundations of an evo-devo computational theory is now a new target horizon in the scientific exploration of speech origins.

## Acknowledgements

This work benefitted from ERC Starting Grant 240007 funding.